\documentclass[11pt,a4paper]{article}
\usepackage{times,latexsym}
\usepackage{url}
\usepackage{placeins}
\usepackage{float}
\usepackage[T1]{fontenc}
\usepackage{graphicx} 

\usepackage{amsmath}
\usepackage[acceptedWithA]{tacl2021v1} 

\usepackage{xspace,mfirstuc,tabulary}

\newif\iftaclinstructions
\taclinstructionsfalse 
\iftaclinstructions
\renewcommand{\confidential}{}
\renewcommand{\anonsubtext}{(No author info supplied here, for consistency with
TACL-submission anonymization requirements)}
\newcommand{\instr}
\fi

\iftaclpubformat 

\else

\fi

\title{Human-like Fleeting Memory Improves Language Learning but Impairs Reading Time Prediction in Transformer Language Models}

\author{
  Abishek Thamma\textsuperscript{1,2} \quad
  Micha Heilbron\textsuperscript{1,3} \\[6pt]
  \textsuperscript{1} University of Amsterdam, Amsterdam Brain and Cognition; Amsterdam, Netherlands  \\
  \textsuperscript{2} Vrije Universiteit Amsterdam, Department of Informatics; Amsterdam, Netherlands \\
 \textsuperscript{3} Max Planck Institute for Psycholinguistics; Nijmegen, Netherlands \\
  {\tt\{a.thamma,m.heilbron\}@uva.nl}
}

\begin{document}

\maketitle
\begin{abstract}
Human memory is fleeting. 
As words are processed, the exact wordforms that make up incoming sentences are rapidly lost. 
Cognitive scientists have long believed that this limitation of working memory may, paradoxically, \emph{help} in learning language – an idea supported by classic connectionist modelling work. 
The rise of Transformers appears to challenge this idea, as these models can learn language effectively, despite lacking working memory limitations or other architectural recency biases. 
Here, we investigate the hypothesized benefit of fleeting memory for language learning in tightly controlled experiments on transformer language models. 
Training transformers with and without fleeting memory on a developmentally realistic training set, we find that fleeting memory consistently improves language learning (as quantified by both overall language modelling performance and targeted syntactic evaluation) but, unexpectedly, impairs surprisal-based prediction of human reading times. 
Interestingly, follow up analyses revealed that this discrepancy -- better language modeling, yet worse reading time prediction -- could not be accounted for by prior explanations of why better language models sometimes fit human reading time worse. 
Together, these results support a benefit of memory limitations on neural network language learning – but not on predicting behavior.  
 \end{abstract}

\maketitle

\section{Introduction}

Human memory is fleeting: as a reader or listener processes language, the exact wordforms that make up incoming sentences are quickly and inescapably lost. 
A popular belief in cognitive science is that this inherent limitation of human memory may paradoxically \emph{help} in language learning \citep{newport1990maturational,christiansen_now-or-never_2016,rowland2013understanding}. 
First, the fleetingness of human memory is thought to impose a recency bias, guiding resources to more relevant, local regularities. 
Second, it is thought to provide an \emph{incentive for abstraction}: because the memory capacity for recent exact wordforms is so limited, the learner would be driven to discover abstractions (such as chunking, linguistic structure) as a means of compression, rather than learn surface-level statistical regularities between words (see \citealp{christiansen_now-or-never_2016} for review).  
\begin{figure}[h!]
    \centering
    \includegraphics[width=0.9\columnwidth]{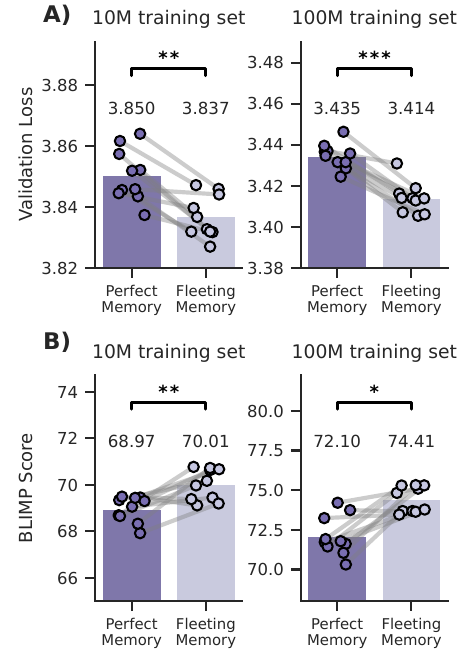}
    \caption{Fleeting memory, combined with an echoic memory buffer, improves language learning, consistently across training runs and training set sizes. 
    \textbf{(A)} Validation loss (lower is better) comparison between models with perfect memory and fleeting memory. 
    \textbf{(B)} BLiMP accuracy (higher is better), a measure of linguistic knowledge, for the same conditions. 
    In all panels, dots with connecting lines are paired training runs with identical weight initialisation and data sampling. Stars indicate significance levels of the pairwise (within-seed) differences (bootstrap t-test against zero): *$p$ < 0.05 (*), *$p$ < 0.01 (**), *$p$ < 0.001 (***).}
    \label{fig:main_language}
\end{figure}
The benefits of memory limitations for learning were famously studied in neural networks in classic connectionist modelling work by \citet{elman1993learning}. 
Using simulations of a toy language and simple recurrent neural network models, Elman found that reducing `working memory' by limiting the input window during training helped the network learn the artificial grammatical structure more effectively. 
Due to Elman's work and that of others (\citealp{christiansen_now-or-never_2016}), the notion that the limitation of memory paradoxically benefits language acquisition became highly influential and can be seen as a cornerstone of the cognitive science of human learning (e.g. \citealp{dehaene2021we,rowland2013understanding}). 

However, the success of the transformer architecture appears to be at odds with this idea. Transformer language models can learn language effectively, including abstract syntactic structure \citep{hu2020systematic,linzen2021syntactic,wilcox2024using,futrell2025linguistics} despite having perfect memory of words within their context window, and lacking any form of inherent recency bias or memory decay, unlike earlier recurrent neural language models. 
Yet, as  the transformer’s success derives from multiple factors including scale, it does not directly inform whether fleeting memory benefits learning, even within modern neural networks, especially in a developmentally realistic data regime.

Here, we investigate the benefit of fleeting memory on language learning in modern neural networks directly, using transformers as a testbed. 
To test for the effect of fleeting memory on language learning in transformers, we propose \emph{fleeting memory transformers}, a simple modification of the transformer architecture that adds a memory decay to the self-attention operation to simulate human-like forgetting. 
By running tightly controlled experiments in which we train models with and without fleeting memory on a developmentally plausible training set (BabyLM; \citealp{warstadt_findings_2023}), we ask whether fleeting memory i) benefits language learning, based on language internal evaluation; ii) renders language models more ‘human-like’, in terms of their ability to predict human reading behaviour. 
At human-scale data, these dimensions typically correlate: better models predict reading times better (\citealp{wilcox_predictive_2020,goodkind_predictive_2018}); moreover, a more human-like architecture might independently better capture human processing.

Our experiments show that, consistently across training runs and network initialisations, adding a human-like memory decay improves both overall language modelling performance, and accuracy on targeted syntactic evaluation – but only when the memory decay is combined with an “echoic memory buffer”, which perfectly retains the initial 3-7 words. 
Intriguingly, however, the models, while consistently better on both language modelling performance and syntactic evaluation, perform worse at surprisal-based prediction of human reading behavior, in a way that cannot be accounted for by prior explanations of why better language models sometimes do worse at reading time prediction.

\begin{figure*}[t!]
    \centering
    \includegraphics[width=0.9\linewidth, trim={1cm 0cm 1cm 0cm}]{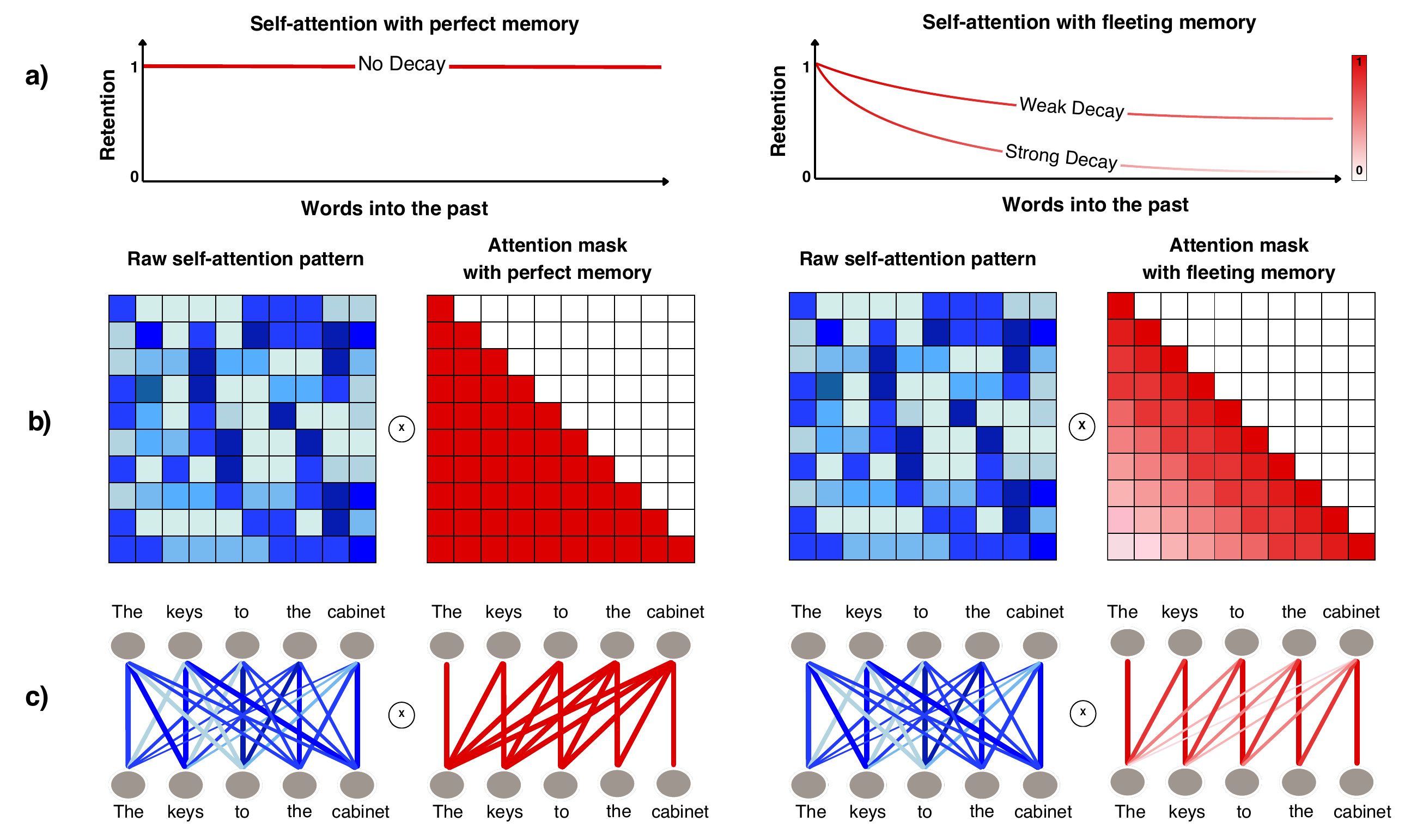}
\caption{(a) Standard Transformers have a retention over entire context window (left), while in our implementation the retention decays as function of distance of current word to the past (Right). \\
		(b) Standard causal self-attention pattern is usually applied to mask future tokens for causal language modeling (Left), while in our implementation an additional bias is added to the causal self-attention mask, where the attention mask fades as function of distance (Right). Shade of red denotes strength of mask. Each word can only attend to itself or preceding words.
	}
    \label{fig:enter-label}
\end{figure*}

\section{Related work}

Studies on the hypothesized benefits of memory limitations for language learning in neural networks go back to the early years of connectionism. 
 In particular, \citet{elman1993learning} reported that systematically increasing the complexity of input to recurrent neural networks improved the models' ability to learn simple toy languages. 
Elman explored this concept of "starting small" in two ways: by imposing resource constraints, in the form of memory limitations that would `simplify' the processing, and secondly by incrementally increasing the complexity of training material itself, in what later became known as \emph{curriculum learning}.

More recently, in the context of transformers, several papers have revisited Elman's ideas of "starting small" as a means to improve learning efficiency in developmentally realistic data ranges in the context of the BabyLM challenge (\citealp[]{warstadt_findings_2023}). 
However, these studies have focused on the curriculum aspect of starting small -- overall not finding consistent benefits of curriculum learning -- and not on memory limitations, as we do here. 
Moreover, note that in Elman's original work, the memory capacity started small but gradually expanded over training. 
By contrast, we consider a fixed memory limitation, simulating the more general effect of limited memory on learning \cite{christiansen_now-or-never_2016,dehaene2021we,rowland2013understanding}; though making memory capacity dynamic  could be an interesting future extension (see \emph{Discussion}). 

Incorporating a recency bias into transformers has also been explored outside the context of cognitively inspired language modeling. 
Most prominently, \citet{press_train_2022} propose ALiBi, a technique that adds linear biases to the attention scores to penalize them proportional to their distance. 
However, the primary goal of their approach is to allow for length generalization during inference time and overcome the constraints posed by traditional positional encoding, not to incorporate a human-like recency bias. 
Additionally, ALiBi's learnable per-head biases (e.g., 36 coefficients in a 6-layer, 6-head setup) results in a potentially quite complicated recency bias, that tends to operate over much larger timescales than human working memory, making it difficult to quantify "memory decay strength" as a single interpretable value.
Rather than competing with such approaches on length generalization, we utilize a different tool for a different purpose: we impose a single, fixed recency bias at human working memory scales, specifically to test the cognitive hypothesis that memory limitations benefit learning \citep{elman1993learning}.

In the cognitive modeling domain, a range of recent works have explored memory limitations or recency biases in language models, motivated by the cognitive implausibility of transformers' perfect verbatim memory \citep{armeni2022characterizing}. 
However, most of these works only explore the effect of memory limitations or recency bias on \emph{inference}, i.e. on top of a pretrained language model without such limitations, rather than on its effect on learning. 
Such memory-based accounts of processing difficulty suggest that humans struggle when retrieving distant or similar items from working memory during language comprehension. 
Various approaches have been used to model these limitations by constraining either memory capacity/width (e.g., \citealp{timkey2023language}, limiting the number of heads), or implementing a decay that creates imperfect memory representations. 
The lossy-context surprisal framework \citep{futrell2020lossy} explores this second approach, proposing that human processing difficulty reflects prediction from such imperfect memory representations. 
Building on this, \citet{hahn2022resource} developed a resource-rational model that optimizes which words to retain in memory, showing that lossy memory better predicts human reading times than perfect memory, particularly for complex recursive structures like center embeddings. However, these models apply memory constraints to already trained language models to explain processing difficulty, rather than examining whether such constraints benefit learning itself.

Formally most similar to our approach, \citet{vaidya_humans_2023} include a power-law decay into self-attention to simulate fleeting memory in pre-trained GPT-2 models. 
However, they too did not focus on learning, but on a very specific form of inference: processing of repeated texts. 
They describe how, for repeating texts, human and language model next-word predictions substantially diverge (as the language model predictions become almost perfect) -- but that a memory decay at inference time reduces this divergence and improves human LM similarity for such repeating texts. 
Recently, and beyond the context of repeated texts, \citet{de_varda_locally_2024} also explored adding a recency bias to pre-trained GPT-2 models.  
They found that a recency bias improved reading time prediction -- but this bias was fitted to optimize reading times directly, and the optimal bias was very small in magnitude.

\citet{clark_linear_2024} extend this approach to training.
Focusing on reading-time prediction, they compared two types of recency bias, namely the one proposed by \citet{de_varda_locally_2024} and ALiBi \cite{press_train_2022}. 
They found that when applied at train-time, the approach of \citet{de_varda_locally_2024} did not lead to measurable improvements in reading time prediction, while ALiBi improved reading time prediction accuracy.
While they interpret this improvement in the light of human memory decay, the exact link seems less direct, because as mentioned, the linear attentional biases in ALiBi are not cognitively inspired, and tend to operate over a different timescale.  
Moreover, the bias is different for different heads, making it difficult to compare the `overall' bias to human memory decay. 

Concurrent work by \citet{mita2025developmentally} revisits Elman's dynamic approach, starting with strong working memory constraints and relaxing them over training until the model concludes with perfect memory. 
While they report improvements on a targeted syntactic benchmark, these emerge primarily late in training as constraints are significantly relaxed, and do not evaluate language modeling loss or behavioral alignment.

In our work, we evaluate the effects on learning of incorporating an analogue of memory decay, implemented using a fixed, non-trainable, cognitively inspired recency bias with interpretable hyperparameters (decay rate and buffer size). 
We evaluate the performance of this approach both in terms of language learning, i.e. performance on overall language modeling and targeted syntactic evaluation, and the ability to predict human behavioral data through reading time fit. 
Critically, all analyses are statistically evaluated and performed using controlled experiments that specifically isolate the effect of fleeting memory, while controlling for stochasticity across training runs. 

\section{Methodology}

\subsection{Architecture}
We study fleeting memory in the context of standard transformer-based autoregressive language model, following the GPT2 architecture \citep{radford_language_2019}. 
We focus on transformers because they lack any inherent locality bias -- attention is uniform over the context window by default  -- making them the cleanest testbed. 
GPT-2 is a vanilla decoder-only transformer, representative of the broader model class. 
The model size was scaled down to match the babyLM dataset size, roughly following scaling trends from \citet{kaplan_scaling_2020}, resulting in a model of 6 layers, 6 attention heads, and a hidden state dimensionality of 384.  
A BPE tokenizer with a vocabulary of 8,000 words was constructed and the models were trained for 44,000 iterations (approximately $\sim$200 epochs over the 10M dataset and $\sim$20 epochs for the 10M), with a batch size of 32. 
The models trained have 13.69M trainable parameters. 

\subsection{Fleeting memory implementation}

We implement fleeting memory as a fixed, non-trainable recency bias applied to the self-attention weights after the softmax operation. This modifies the standard attention mechanism as follows:
\begin{equation}
    \label{eq:fm_attention}
    \resizebox{0.89\linewidth}{!}{$\text{Attn}_{\text{FM}}(Q,K,V) = \left( \text{softmax}\left( \frac{QK^T}{\sqrt{d_k}} \right) \odot B \right) V$}
\end{equation}
Note that after masking, attention weights do not sum to one; however, the subsequent Layer Normalization is scale-invariant, so only the relative attention pattern affects learning. The bias matrix $B$ contains retention values between 0 and 1, which decay as a function of token distance $d$. Motivated by models of forgetting in cognitive science \cite{donkin_power-law_2012, lin_critical_2017}, we use a power-law retention function:
\begin{equation}
B(d) = 
\begin{cases} 
  1 & \text{if } d < E \\
  1 - \left( \frac{d - E + 1}{n - E} \right)^{\frac{1}{e\alpha}} & \text{if } E \le d < n
\end{cases}
\label{eq:powerlaw_decay}
\end{equation}
This piecewise function includes an ``echoic memory'' buffer -- an initial period of size $E$ where retention is perfect. For subsequent tokens ($d \ge E$), retention decays as a power-law over the remainder of the context window, $n$. The steepness of this decay is controlled by the hyperparameter $\alpha$, and $e$ is Euler's number.

\subsection{Dataset}
In order to train the models in a human-like data quantity and register, models were trained on the training set of the BabyLM challenge \citep{warstadt_findings_2023}. 
For our primary experiments, the 10M dataset was used, which corresponds to the amount of input received by children in their first 2-5 years. 
Subsequently, we extended our analysis to the 100M dataset, which corresponds to the rough equivalent of the input received by age 12. The dataset is heterogeneous, containing a variety of data sources such as CHILDES \cite{macwhinney_childes_2004}, OpenSubtitles \cite{lison_opensubtitles2016_2016}, children's stories, and others, with approximately 56\% being transcribed speech and about 40\% being child-directed or child-appropriate language.

\subsection{BLiMP} 

BLiMP is a benchmark for linguistic minimal pair judgments \cite{warstadt_blimp_2019}, containing 67 datasets of 1,000 minimal pairs across 12 linguistic phenomena. Models are evaluated on whether they assign a higher probability to the acceptable sentence in each pair, analogous to human acceptability judgments. Scores represent the proportion of correct judgments per subtask, with the overall score calculated as an unweighted average across all subtasks following the BabyLM evaluation pipeline.

\subsection{Reading Time Analysis}

We evaluate the models' fit to human reading times using linear mixed-effects regression models. We compare a baseline model against a full model that incorporates LM surprisal along with the baseline predictors; the resulting difference in Log-Likelihood ($\Delta$LL) quantifies how much better the model's predictions explain human processing delays beyond what these lexical baselines already capture. This approach isolates the unique contribution of the language model to processing difficulty.

Our primary analysis employs a rigorous baseline specification. 
We model fixed effects for low-level lexical features (word length and unigram frequency from \citealp{brysbaert_moving_2009}) for the current word, as well as for the two preceding words to account for spillover effects. For the Dundee corpus, we additionally control for whether the previous word was skipped. Crucially, we include by-subject random slopes for all fixed effects (both current and lag features) to ensure conservative statistical estimates. These models were fit using \texttt{lme4} \citep{bates_lme4} on data from Natural Stories (self-paced reading times from 181 participants reading 10 stories, 10,256 words) and the Dundee Corpus (gaze durations from 10 subjects reading 20 news articles, 51,500 words).

Finally, to ensure our results are robust and not driven by a specific choice of covariates, we fitted an alternative ``item effects'' model \citep{oh_why_2023, oh2024frequency}. This specification replaces the explicit lexical covariates (length, frequency, and their spillovers) with a random intercept for each unique word type, directly factoring out item-specific variance.

\subsubsection*{Frequency-based Error Analysis}
To investigate whether impairment in reading time prediction could be explained by memorization as proposed for very large language models \citep{oh_why_2023, oh2024frequency}, we conducted a follow-up analysis that assessed the prediction performance as a function of word frequency. 
To do so, we first partitioned the data from each corpus into five quintiles based on log-frequency and calculated the Mean Squared Error (MSE) of the regression residuals for each quintile. 
To probe for the second signature of memorization -- systematic underprediction of reading times for rare words -- we subsequently separated the data points within each quintile into underpredicted and overpredicted instances. 
For this step, following \citet{oh2024frequency}, we computed the Sum of Squared Errors (SSE) to quantify total error, because the number of under- and overpredicted items can differ across models. 

\subsection{Statistical Testing}
When analyzing model performance for the evaluation measures, multiple seeds for each configuration (10 seeds) were tested. 
This was done in order to correct for stochasticity in training and also analyze consistency across seeds, where each trained model is treated as a separate `participant'.
We define a model `participant' by its random seed, which determines both the initial weight configuration and the data sampling order (batch shuffling). 
By fixing the seed across conditions, we ensure paired comparisons are ceteris paribus: the models differ only in their memory architecture, while the initialization and training data sequence remain identical (up to GPU floating-point precision).
Statistical testing was performed across participants (seeds) and since the number of participants was low, data-driven bootstrap t-tests were used. These involve resampling a null-distribution with zero mean (by removing the mean), counting across bootstraps how likely a t-value at least as extreme as the true t-value was to occur. Each test used at least $10^4$ bootstraps; p values were computed without assuming symmetry (equal-tail bootstrap; \citealp{rousselet2023introduction}). Confidence intervals (in the figures and text) were also based on bootstrapping.

\subsection{Code}

Models were built using custom code in PyTorch, built on top of nanoGPT \cite{Karpathy2022}. 
For the BLiMP evaluation, we used the evaluation pipeline created by the organizers of
the BabyLM challenge. All models were trained on A100 GPUs on Snellius supercomputing cluster. Code available at \url{https://github.com/drhanjones/fmt-llm}

\section{Results}

\subsection*{Naive fleeting memory impairs language learning} 

\begin{figure}[h]
    \centering\includegraphics[width=\linewidth]{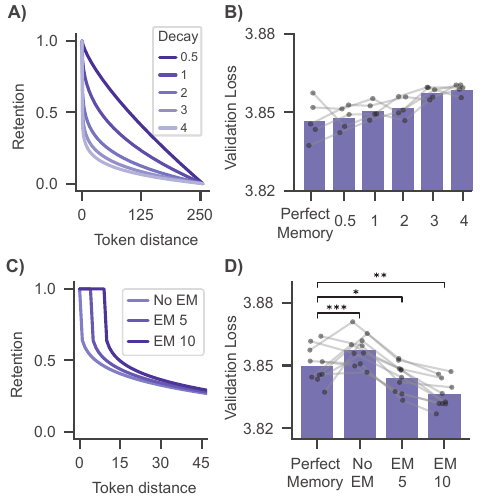}
    \caption{ \textbf{A)} Naive fleeting memory. Retention values with respect to token recency, for different decay rates. 
    \textbf{B)} Effect of naive fleeting memory on cross-entropy loss across five seeds. 
    \textbf{C)} Fleeting memory with and without echoic memory, illustrated for a decay rate of 2. 
    \textbf{D)} Effect of echoic memory on cross-entropy loss, for a decay rate of 2, versus perfect memory (no decay).  
    For the consistency across decay strengths and EM-buffers, see Figure \ref{fig:supp_em-decay}.
    In all panels, dots with connecting lines represent individual (pairwise) training runs for different conditions. 
    Stars indicate significance-levels of the pairwise (within-seed) difference, across different runs (bootstrap t-test against zero): $p < 0.05$ (*), $p < 0.01$ (**), $p < 0.001$ (***).
    }
    \label{fig:main_echoic_naive}
\end{figure}

We first evaluated a "naive" implementation of fleeting memory,
i.e. a memory retention function that immediately starts decaying from the first token into the past (see Fig \ref{fig:main_echoic_naive}a). 
We assessed the effect of such a memory decay on overall language modeling performance, as quantified via the final loss on the validation dataset from the BabyLM challenge, for a range of memory decay strengths. 
To isolate the effect of memory decay from other factors contributing to validation loss variability, we trained a range of models with different random seeds (affecting  weight initialization and data sampling) for every decay value, but shared random seeds across decays, to allow for paired comparisons (see Fig \ref{fig:main_echoic_naive}b and \emph{Methods}).

Interestingly, this revealed that naive fleeting memory \emph{impairs} language model performance: validation loss was lowest for models without memory decay (i.e. perfect memory models) and increased slightly but monotonically as a function of decay strength, highly consistently across training run seeds (Figure \ref{fig:main_echoic_naive}b).

\subsection*{Fleeting memory improves language learning when combined with echoic memory buffer}

At face value, the initial result challenges the hypothesis that memory limitations can improve language learning. 
However, qualitative inspection of the models suggested that the naive fleeting memory decay introduced too strong a decay.
In particular, we observed from model completions that naive fleeting memory introduced many spelling errors, suggesting even the most local (within-word, across-token) dependencies were disrupted. 
This stands in contrast with human memory, which, while fundamentally fleeting, retains the most recent past near-perfectly due to a brief sensory buffer period (`echoic' or `iconic' memory; \citealp{baddeley1992working}).
To account for this, we explored a modified version of fleeting memory, in which the memory decay is preceded by an echoic memory buffer: an initial period of 5-10 tokens ($\pm$ 3-7 words) during which the retention is perfect (Fig \ref{fig:main_echoic_naive}C; see \citealp{harrison2020ppm} for a similar approach in music modeling). 

When we evaluated these models, in a larger experiment (10 seeds per condition), we first confirmed that naive fleeting memory impairs language modelling performance, and that this effect was statistically significant (for a decay of 3: mean $\Delta L = +0.0073$, 95\% CI $[+0.0044, +0.0103]$, bootstrap $t$-test: $p = 0.0008$; see Figure \ref{fig:main_echoic_naive}D). 
Strikingly, however, for fleeting memory models with an echoic memory buffer, we observed the opposite pattern: a lower validation loss compared to perfect memory models, both for echoic memory of 5 (mean $\Delta L = -0.0059$, 95\% CI $[-0.0089, -0.0028]$, bootstrap $t$-test: $p = 0.0138$) and 10 (mean $\Delta L = -0.0135$, 95\% CI $[-0.0177, -0.0090]$, bootstrap $t$-test: $p = 0.0012$).
To confirm that these results were not an artifact of a specific hyperparameter choice, the analysis was repeated across 12 distinct combinations of decay strength and echoic memory buffer size (120 training runs in total). 
This confirmed that naive fleeting memory (no buffer) consistently impaired performance, while the combination of an echoic buffer and sufficiently strong decay consistently improved it (Figure \ref{fig:supp_em-decay}).

To further evaluate the robustness of this effect, we repeated the experiment on the larger 100M BabyLM training set, finding the same effect: fleeting memory improves language modelling performance (reduces loss), and this effect is highly consistent across training runs (mean $\Delta L = -0.0207$, 95\% CI $[-0.0237, -0.0169]$, bootstrap $t$-test: $p < 0.0001$; see Figure \ref{fig:main_language}A). For the subsequent evaluations, we focus on a representative model configuration with an echoic memory buffer of 10 tokens (approximating human capacity) and an intermediate decay rate of $\alpha=2$. Note, however, that the general benefits of fleeting memory are robust across a wide range of hyperparameter settings (see Figure \ref{fig:supp_em-decay} )

\subsection*{Fleeting memory improves linguistic knowledge}

We then evaluated the effect of fleeting memory not just on next-word prediction ability (cross-entropy loss) but also on syntactic knowledge, using targeted syntactic evaluation on the benchmark of linguistic minimal pairs (BLIMP; \citealp{warstadt_blimp_2019}). 
Indeed, we observed that fleeting memory models showed improved syntactic knowledge -- and this was found consistently for both models trained on the 10M training set (mean $\Delta_{acc} = 1.04\%$, 95\% CI $[0.52\%, 1.54\%]$, bootstrap $t$-test: $p = 0.0086$) and those on the 100M training set (mean $\Delta_{acc} = 2.30\%$, 95\% CI $[1.35\%, 3.17\%]$, bootstrap $t$-test: $p = 0.0102$; see Figure \ref{fig:main_language}B). 
This suggests that the benefits of fleeting memory extend beyond simple next-token prediction to more abstract and theoretically significant aspects of language processing.

\subsection*{Differential impact of fleeting memory across linguistic subtasks}

When we subdivided the aggregate BLiMP scores across the twelve broad linguistic phenomena (see appendix; Figure \ref{fig:blimp_10M} and \ref{fig:blimp_100M}), we observed that the improvement conferred by fleeting memory is not uniform, but was driven by a specific subset of phenomena. 
Notably, consistent and significant gains were evident in tasks such as Subject-Verb Agreement, Anaphor Agreement, and Argument Structure, which rely on relatively local syntactic context and dependencies, where a fleeting memory bias would theoretically be advantageous.

Critically, fleeting memory did not significantly impair accuracy on any of the phenomena in BLiMP. 
Nevertheless, several phenomena exhibited no improvement.
Indeed, for both the 10M and 100M models, "Irregular Forms" remained unaffected, as anticipated, as this is a test of lexical knowledge rather than dependencies between tokens. "Determiner Noun Agreement" also showed consistently no improvements -- which is expected, since it involves among the most local regularities that fall within the echoic memory buffer.

\subsection*{Fleeting memory impairs reading time prediction}

\begin{figure}[!h]
    \centering
    \includegraphics[width=\columnwidth]{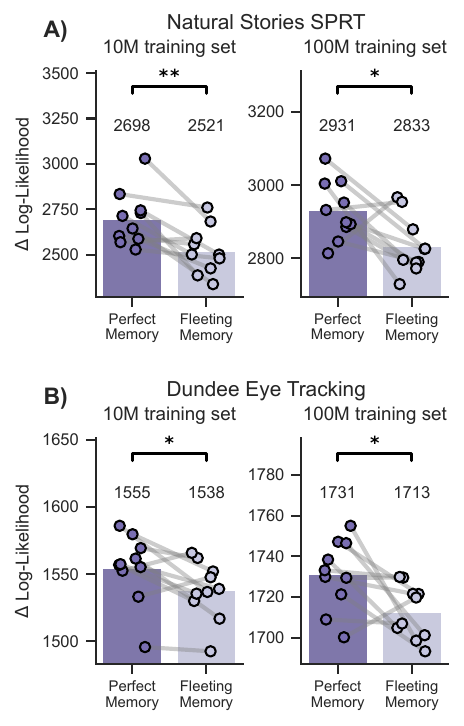}
    \caption{
    Fleeting memory impairs surprisal-based prediction of human reading times, across different datasets and training set sizes. 
    \textbf{(A)} $\Delta$ Log-Likelihood (higher is better) for predicting self-paced reading times on the Natural Stories corpus. 
    \textbf{(B)} $\Delta$ Log-Likelihood (higher is better) for predicting gaze durations on the Dundee eye-tracking corpus. 
    In all panels, dots with connecting lines are paired training runs with identical weight initialization and data sampling. 
    Stars indicate significance levels of the pairwise (within-seed) differences (bootstrap t-test against zero): \textit{ns} (not significant), *$p$ < 0.05 (*), *$p$ < 0.01 (**).
}
    \label{fig:main_rt}

\end{figure}

The improvements in language modeling and syntactic knowledge raised a natural question: would fleeting memory models also better capture human language processing? 
We tested this hypothesis by assessing behavioral alignment in the form of surprisal-based reading time prediction. 
We first assessed self-paced reading times from the Natural Stories corpus \citep{futrell2021natural}. 
Strikingly, we found that fleeting memory significantly \emph{impaired} reading time prediction. 
This was robust across training scales, with models trained on the 10M dataset showing a decrease in delta-log-likelihood of $\Delta LL = -176.29$ (95\% CI $[-253.67, -95.93]$, bootstrap $t$-test: $p = 0.0033$) and models trained on the 100M dataset showing the same pattern (mean $\Delta LL = -98.13$, 95\% CI $[-165.43, -22.99]$, bootstrap $t$-test: $p = 0.0365$; see Figure \ref{fig:main_rt}A).

We considered whether this might be an artifact of the self-paced reading paradigm, where reading times are assessed through button presses, which is not very naturalistic. 
Overall, the same pattern was found: for both training sets, fleeting memory induced a quantitative impairment on the delta log likelihoods (10M: mean $\Delta LL = -16.92$, 95\% CI $[-29.89, -4.34]$, bootstrap $t$-test: $p = 0.02$, 100M: mean $\Delta LL = -18.30$, 95\% CI $[-32.41, -5.00]$, bootstrap $t$-test: $p = 0.01$; Figure \ref{fig:main_rt}B). 
Crucially, we obtain equivalent results using the alternative ``item effects'' regression model (see Appendix; Figure \ref{fig:supplementary_rt_itemintercept}), confirming that our findings are robust to regression model specification choices.

This establishes a curious dissociation: fleeting memory enhances objective language modeling performance while simultaneously degrading the models' ability to predict human processing patterns.

\subsection*{The impairment in reading time prediction cannot be fully accounted for by prior explanations}

The finding that fleeting memory improves language model quality while impairing reading time prediction appears paradoxical, but echoes a broader pattern in the literature. 
Early work established that better language models predict human reading times better \cite{goodkind_predictive_2018,wilcox_predictive_2020}, but recent studies have documented a reversal: superior models often exhibit degraded reading time prediction \cite{oh_comparison_2022,shain2024large}, spawning several explanatory hypotheses \cite{oh_why_2023,oh2023twobillion,oh2024frequency,aoyama2025language}. 
We examined whether these prior accounts could explain our results.

The \emph{scale hypothesis} attributes this inversion to superhuman training data quantities, proposing that the positive relationship breaks down around 2 billion tokens—far exceeding human linguistic exposure \cite{oh2023twobillion}. However, this cannot explain our results: we observe the effect with human-scale training data, at just 10M words.

A second influential explanation is the \emph{memorization hypothesis}. 
This starts from the observation that the impairment in reading time prediction is not uniform but is concentrated at specific words, in particular specific open-class words such as proper nouns and named entities \cite{oh_why_2023}, and low-frequency items more generally \cite{oh2024frequency}. 
The reasoning here is that the best models show human-unlike memorization, allowing them to predict infrequent words too well. 
This account predicts two empirical signatures: (i) maximal prediction degradation for infrequent words, and (ii) systematic reading time underprediction, as memorization drives surprisal values for very rare tokens unrealistically low
\citep{oh_why_2023,oh2024frequency}. 

Testing these predictions reveals an intriguing mismatch. 
We observe the first signature: across both corpora and model scales, fleeting memory's impairment concentrates on low-frequency items (see appendix; Figure \ref{fig:supp_quint_overall}). 
However, we find no evidence that the worse reading time prediction of low-frequency words is driven by the underprediction of reading times for low-frequency words, the key signature of memorization-based accounts (Figure \ref{fig:supp_over-under_natural},\ref{fig:supp_over-under_dunee}). 
This indicates that while frequency appears to mediate the effect, the mechanism differs from earlier explanations based on superhuman memorization of low-frequency items. 

\section{Discussion}

Our experiments demonstrate that memory limitations can benefit language learning in transformer language models, supporting theories from cognitive science. 
When equipped with an echoic memory buffer, fleeting memory models consistently outperformed perfect memory controls on both overall language modeling and targeted syntactic evaluation, on a developmentally realistic training set. 
However, these same models exhibited impaired reading time prediction. 
The impairment cannot be attributed to superhuman-scale training data or memorization mechanisms identified in prior work as causes of degraded reading time fit for better language models, suggesting that multiple distinct mechanisms can cause a dissociation between language model quality and reading time prediction accuracy. 

Before discussing theoretical implications, it is important to note the magnitude of the observed effects. 
The reduction in cross-entropy loss is subtle -- roughly the same magnitude as the run-to-run variability from different random seeds. 
However, our methodology, which isolates the effect of fleeting memory across paired training runs, demonstrates that the benefit is consistent. 
Moreover, this small but reliable advantage transfers to a more pronounced and meaningful improvement on BLiMP. 
This consistency and transferability validate the effect as a genuine consequence of memory decay, justifying further interpretation.

A paradoxical aspect of these results is that we find that memory decay improves learning in \emph{transformers} -- the very architecture famous for lacking the recency biases characteristic of previously dominant RNNs. 
Yet we demonstrate that reintroducing memory limitations enhances both language modeling performance and syntactic knowledge acquisition. 
However, we should stress that we do not believe the results to reflect something about transformer language models in general. 
We focus specifically at the human-scale data regime (10-100M tokens). 
In this data-limited regime, memory decay provides a useful inductive bias: It encodes the statistical fact that most linguistic dependencies are local \cite{futrell_dependency_2020}.
The benefits would likely diminish at contemporary pretraining scales (billions to trillions of tokens), where models can discover this pattern from data alone. Pinpointing the exact data scale at which benefits disappear—and whether model size plays an independent role—remains open, though computationally expensive to determine and orthogonal to our focus on developmentally plausible regimes. 
Moreover, our results may be specific to the developmental corpus we used; texts with genuinely long-range dependencies -- academic papers, novels, code, etc. -- might show different patterns.

What mechanisms underlie the benefits of fleeting memory? The most parsimonious explanation is statistical: memory decay serves as a beneficial inductive bias. While a standard Transformer could in principle learn to attend preferentially to local context, discovering this pattern requires data. Fleeting memory encodes this prior directly, improving sample efficiency -- a computational instantiation of the `less is more' hypothesis \citep{newport1990maturational,elman1993learning}. 
A richer possibility is that fleeting memory creates an incentive for abstraction: because exact wordforms decay rapidly, the learner must discover higher-level abstractions to preserve information \citep{christiansen_now-or-never_2016}. 
The pattern of BLiMP improvements can be seen as in line with this abstraction hypothesis: fleeting memory helps in learning syntactic phenomena requiring structural analysis. 
However, our current findings are not sufficient to conclude whether fleeting memory operates only through statistical biasing or (also) drives qualitatively different learning. 
Future work could try to do so with more mechanistic analyses -- for instance by examining layer-wise abstraction, attention head specialization, or the temporal dynamics of feature emergence \citep{aoyama2025language}. 

While we observe that fleeting memory hurts reading time prediction, two recent studies reported the opposite. 
\citet{de_varda_locally_2024} found that a recency bias improved behavioral fit when added to pretrained GPT-2; \citet{clark_linear_2024} observed benefits with ALiBi.
This discrepancy probably stems from methodological differences.
\citet{de_varda_locally_2024} applied decay post-hoc to pretrained models and fit decay parameters directly to optimize reading times, thereby potentially limiting theoretical conclusions.
\citet{clark_linear_2024} failed to replicate this result with the decay functions of \citet{de_varda_locally_2024}, obtaining improvements only with ALiBi -- but ALiBi's head-specific, trainable biases spanning many tokens are more difficult to compare to human memory constraints.
By contrast, we impose an interpretable, fixed decay during training, and find consistent improvements in language modeling and BLiMP accuracy alongside impairments in reading time prediction.
The consistency of these effects suggests memory decay (applied over training) affects language models differently than these prior works anticipated. 
One might object that our retention function was not entirely fixed -- specifically, the echoic buffer was added after observing poor performance of the naive decay (Figure \ref{fig:main_echoic_naive}). However, unlike approaches that optimize decay parameters directly against language modelling performance or reading time fit, our echoic buffer was motivated by human memory, was never trainable end-to-end, and was based on the 10M dataset alone. 
All effects on BLiMP, reading times, and generalization to the 100M dataset emerged downstream from this theoretically motivated, fixed constraint.

Our results exemplify an emerging paradox in computational psycholinguistics: superior language models often predict human reading behavior worse \citep{oh_why_2023,oh_comparison_2022} (see also \citealp{oh2025model}).  
Our models are not fully in this regime -- which has been documented for models trained on billions of tokens -- as our 100M models predict reading times better than 10M models (especially for Dundee, Figure \ref{fig:main_rt}B), indicating that model quality still correlates positively with behavioral fit at our data scales. 
Yet within each training scale, fleeting memory creates this dissociation: improved language models alongside degraded reading time prediction.
Prior explanations -- based on superhuman scale \citep{oh2023twobillion} or memorization of low-frequency words \citep{oh2024frequency} -- cannot account for our findings. This suggests that multiple mechanisms can produce the competence-alignment dissociation. 

The power-law decay we use is interpretable and not fitted on the data directly, but only represents a crude approximation of human memory. 
Future work could extend it in at least two dimensions. 
First, human forgetting is content-sensitive and non-monotonic. 
Some words persist despite temporal distance, while others vanish quickly. 
Content-dependent retention functions \citep[as in][]{hahn2022resource} could preserve high-information words while letting predictable ones fade, better approximating how humans selectively retain meaningful content. 
Second, human retention changes across development.
Children's working memory expands with age, altering what linguistic patterns they can acquire \citep{gathercole1992phonological}.
Following \citet{elman1993learning}, models could begin with severe memory constraints that gradually relax.
Indeed, concurrent work by \citet{mita2025developmentally} demonstrates that such a memory-based curriculum can benefit syntactic abilities, although their models conclude training with perfect memory. A natural extension would be to combine such dynamic relaxation with a more human-like endpoint, starting severely constrained and relaxing to fleeting memory.

Ultimately, our findings lend new support to the classic view that cognitive limitations can serve as a beneficial inductive bias for learning. 
Yet, the paradoxical impairment in reading time prediction reveals a surprising distinction: a model that is more human-like in its architectural constraints is not necessarily more human-like in its prediction of human behavior. 

\section*{Acknowledgments}
We thank the anonymous reviewers and the action editor Dilek Hakkani-Tur for their constructive feedback, which has greatly helped improve this work. 
This work has been partly funded by the Dutch Research Council (NWO) under Veni grant VI.Veni.231G.043 to MH. 

\raggedbottom
\bibliography{tacl2021}
\bibliographystyle{acl_natbib}

\appendix
\section{Appendix}
\label{sec:appendix}
\subsection*{Supplementary results}
\begin{figure*}[] 
    \centering 
    
    \begin{minipage}[t]{0.46\linewidth}
        \centering
        \includegraphics[width=0.9\linewidth]{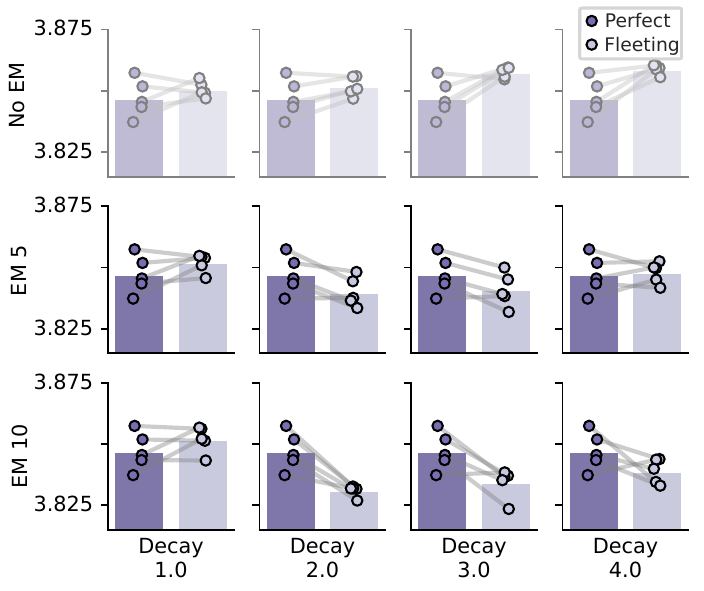}
        \caption{Validation loss on 10M dataset, across 12 decay-strength - buffer-size combinations (120 training runs). Without echoic memory (No EM or ``naive''), decay impairs performance. At sufficient decay strengths and buffer sizes, fleeting memory consistently improves performance.} 
        \label{fig:supp_em-decay}
    \end{minipage}
    \hfill 
    \begin{minipage}[t]{0.46\linewidth}
        \centering
        \includegraphics[width=0.9\linewidth]{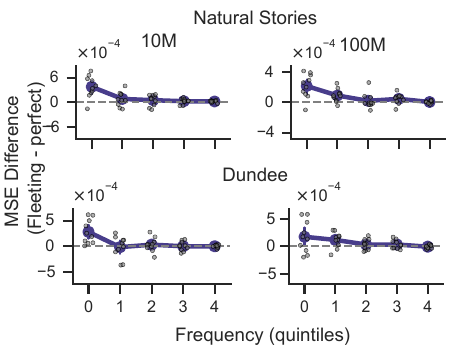}
        \caption{Quintile analysis reveals the effect concentrates on low-frequency words. Difference in reading time prediction error, for both Natural Stories and Dundee for both training set sizes, across different words frequency in the reading time datasets. Small dots indicate paired (within seed) difference in MSE (fleeting - perfect), showing that the fleeting memory models are especially worse at low-frequency words.}
        \label{fig:supp_quint_overall}
    \end{minipage}

    \vspace{1cm}

    \begin{minipage}[t]{0.46\linewidth}
        \centering
        \includegraphics[width=0.9\linewidth]{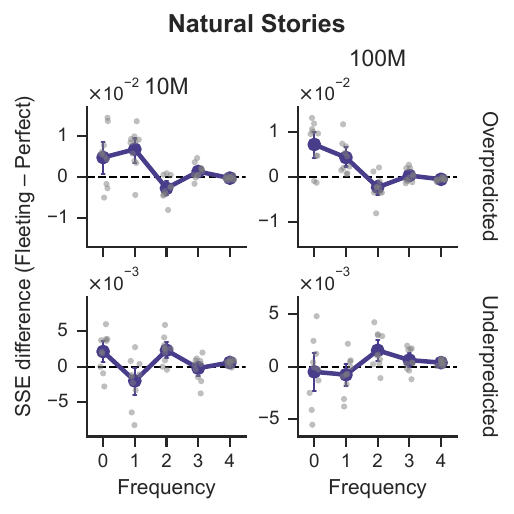}
        \caption{Reading time over/under-predictions as a function of frequency, for Natural Stories Corpus. Similar to Figure \ref{fig:supp_quint_overall}, but calculating the difference in total error for words of which the reading time was unpredicted and over-predicted separately. The SSE difference is normalized within each condition (quintile and prediction type) relative to the average 'Perfect' model error to allow for comparison across scales. 
        We do not observe that the elevated difference in residual error at low-frequency words is driven specifically by \emph{underpredictions} of reading times, as observed by \citet{oh2024frequency}.} 
        \label{fig:supp_over-under_natural}
    \end{minipage}
    \hfill 
    \begin{minipage}[t]{0.46\linewidth}
        \centering
        \includegraphics[width=0.9\linewidth]{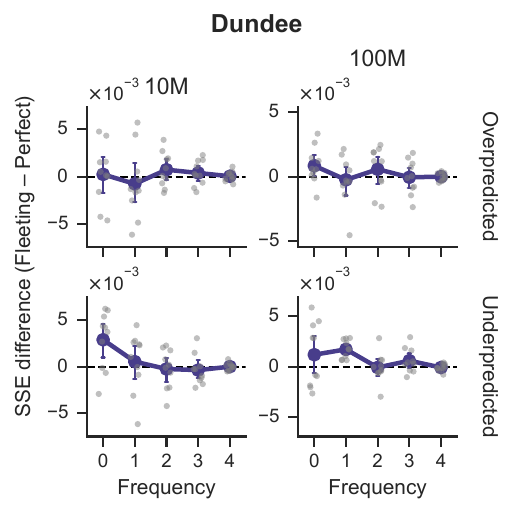}
        \caption{Same as Figure \ref{fig:supp_over-under_natural}, but for Dundee corpus. Here too, we do not observe clearly that the elevated difference in residuals at low-frequency words is driven specifically by \emph{underpredictions} of reading times, as in \citet{oh2024frequency}.} 
        \label{fig:supp_over-under_dunee}
    \end{minipage}

\end{figure*}

\begin{figure*}
    \centering
    \includegraphics[width=.95\linewidth]{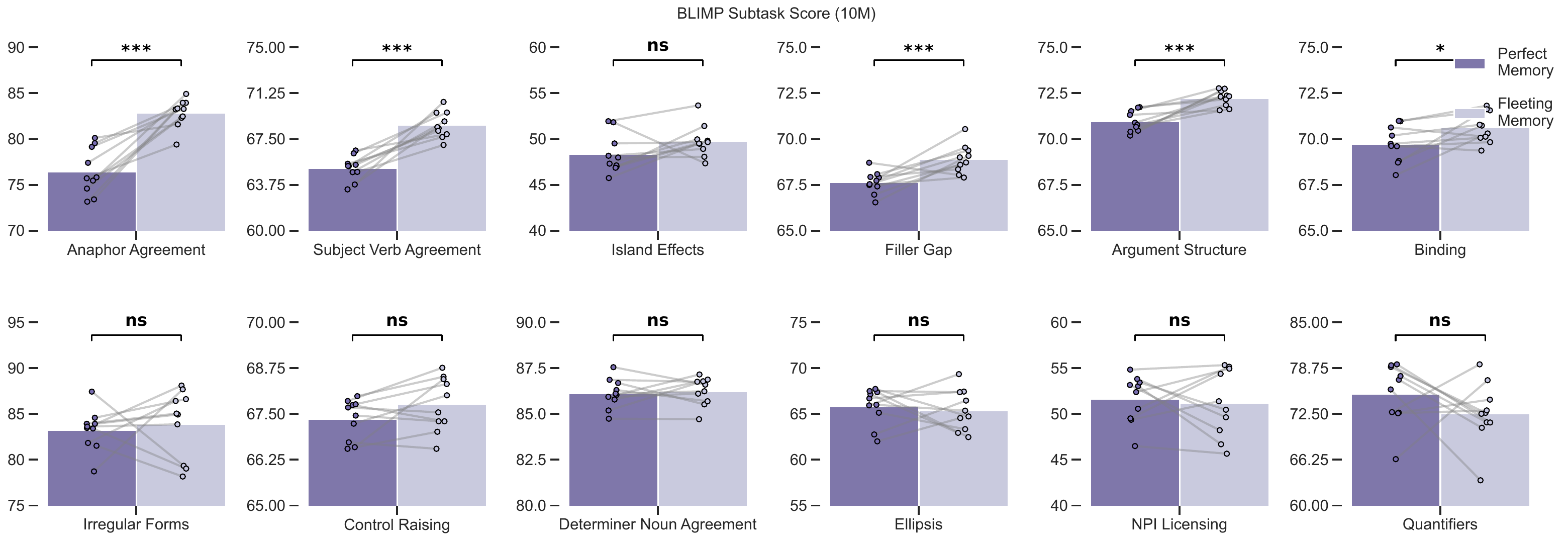}
    \caption{10M models (with and without fleeting memory) performance on BLiMP accuracy on subtasks across 12 broad linguistic phenomena, sorted  by the average improvement numerical by fleeting memory (highest first). } 
    \label{fig:blimp_10M}
\end{figure*}

\begin{figure*}
    \centering
    \includegraphics[width=0.95\linewidth]{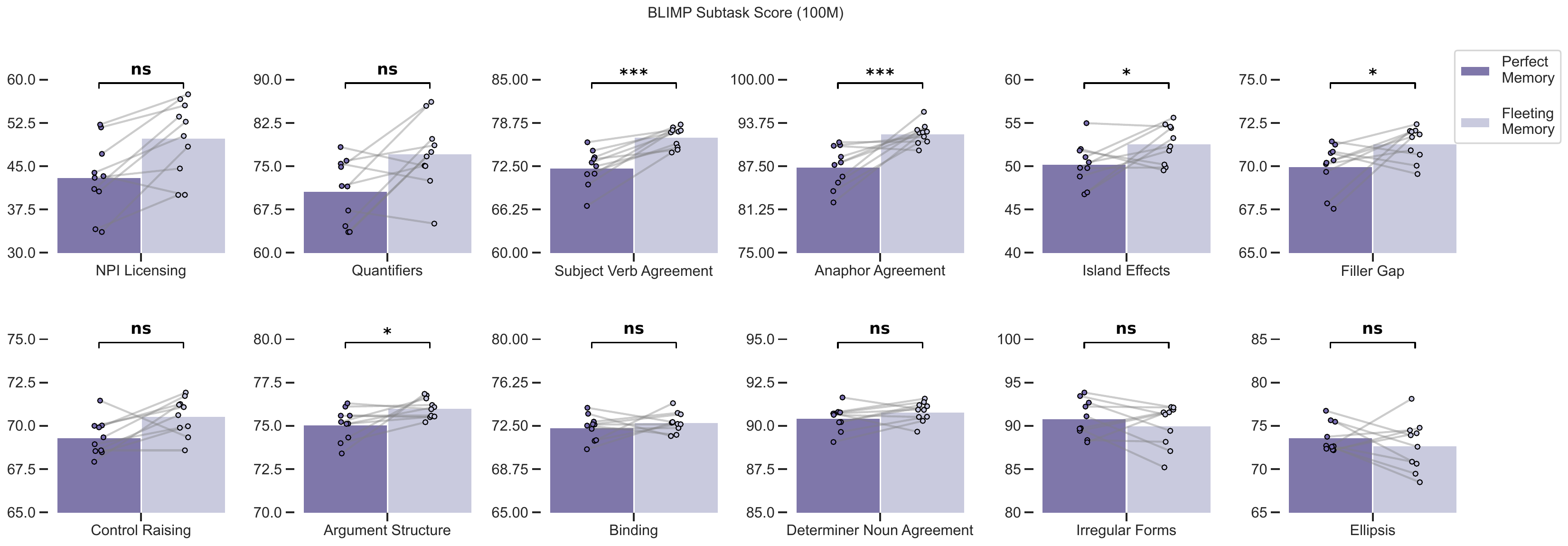}
    \caption{100M models (with and without fleeting memory) performance on BLiMP accuracy on subtasks across 12 broad linguistic phenomena, sorted  by the average improvement numerical by fleeting memory (highest first). } 
    \label{fig:blimp_100M}
\end{figure*}

\begin{figure}[!t]
    \centering
    \includegraphics[width = 0.9\columnwidth]{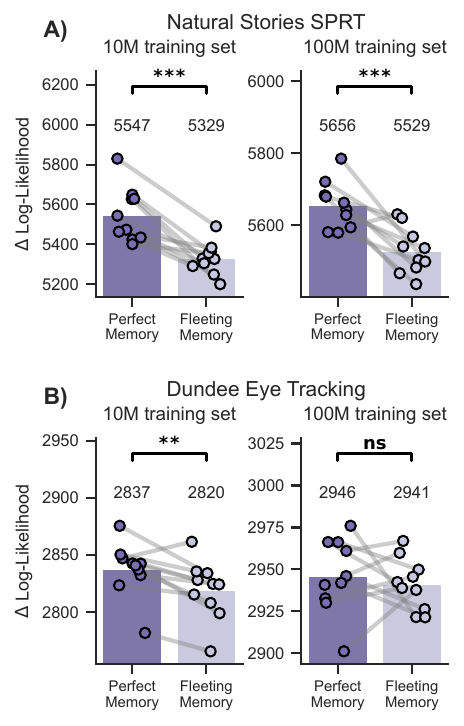}
    \caption{
    Suprisal based reading time prediction modelled using unique random intercepts for each word-type (as opposed to explicit lexical covariates) shows qualitatively identical results.
    \textbf{(A)} $\Delta$ Log-Likelihood (higher is better) for predicting self-paced reading times on the Natural Stories corpus. 
    \textbf{(B)} $\Delta$ Log-Likelihood (higher is better) for predicting gaze durations on the Dundee eye-tracking corpus. 
    In all panels, dots with connecting lines are paired training runs with identical weight initialization and data sampling. 
    Stars indicate significance levels of the pairwise (within-seed) differences (bootstrap t-test against zero): \textit{ns} (not significant), *$p$ < 0.05 (*), *$p$ < 0.01 (**).
    }
    \label{fig:supplementary_rt_itemintercept}
\end{figure}

\end{document}